\documentclass{article}
\usepackage[utf8]{inputenc}
\usepackage{graphicx} 
\usepackage{amsmath}
\usepackage{hyperref}

\usepackage{authblk}
\title{Synthesising Dynamic Textures using Convolutional Neural Networks}
\date{\vspace{-5ex}}
\author[$\dagger$, 1, 2, 3]{Christina M. Funke\thanks{Corresponding Author: christina.funke@bethgelab.org}}
\author[$\dagger$, 1, 2, 4]{Leon A. Gatys}
\author[1, 2, 5]{Alexander S. Ecker}
\author[1, 2, 3, 6]{Matthias Bethge}

\affil[$\dagger$]{contributed equally}
\affil[1]{Werner Reichardt Center for Integrative Neuroscience, University of T\"{u}bingen, Germany}
\affil[2]{Bernstein Center for Computational Neuroscience, T\"{u}bingen}
\affil[3]{Max Planck Institute for Biological Cybernetics, T\"{u}bingen}
\affil[4]{Graduate School of Neural Information Processing, University of T\"{u}bingen, Germany}
\affil[5]{Department of Neuroscience, Baylor College of Medicine, Houston, TX, USA}
\affil[6]{Institute for Theoretical Physics, University of T\"{u}bingen, Germany}

\newcommand{\bx}{\textbf{x}}
\newcommand{\bX}{\textbf{X}}

\newcommand{\bF}{\textbf{F}}
\newcommand{\bG}{\textbf{G}}
\newcommand{\dt}{$\Delta t$ }
\newcommand{\dmt}{\Delta t}

\begin{document}

\maketitle

\begin{abstract}
Here we present a parametric model for dynamic textures. The model is based on spatiotemporal summary statistics computed from the feature representations of a Convolutional Neural Network (CNN) trained on object recognition. We demonstrate how the model can be used to synthesise new samples of dynamic textures and to predict motion in simple movies.
\end{abstract}

\section{Introduction}
Dynamic or video textures are movies that are stationary both in space and time. Common examples are movies of flame patterns in a fire or waves in the ocean. There exists a long history in synthesising dynamic textures (e.g. \cite{doretto_dynamic_2003,kwatra_graphcut_2003,schodl_video_2000,szummer_temporal_1996,wang_generative_2002,wei_fast_2000,xie_synthesizing_2016}) and recently spatio-temporal Convolutional Neural Networks (CNNs) were proposed to generate samples of dynamic textures \cite{zhu_dynamic_2016}. In this note we introduce a much simpler approach based on feature spaces of a CNN trained on object recognition \cite{simonyan_very_2014}. We demonstrate that our model leads to comparable synthesis results without the need to train a separate network for every input texture.

\section{Dynamic texture model}

\begin{figure}
	\center
    \includegraphics[width=0.3\linewidth]{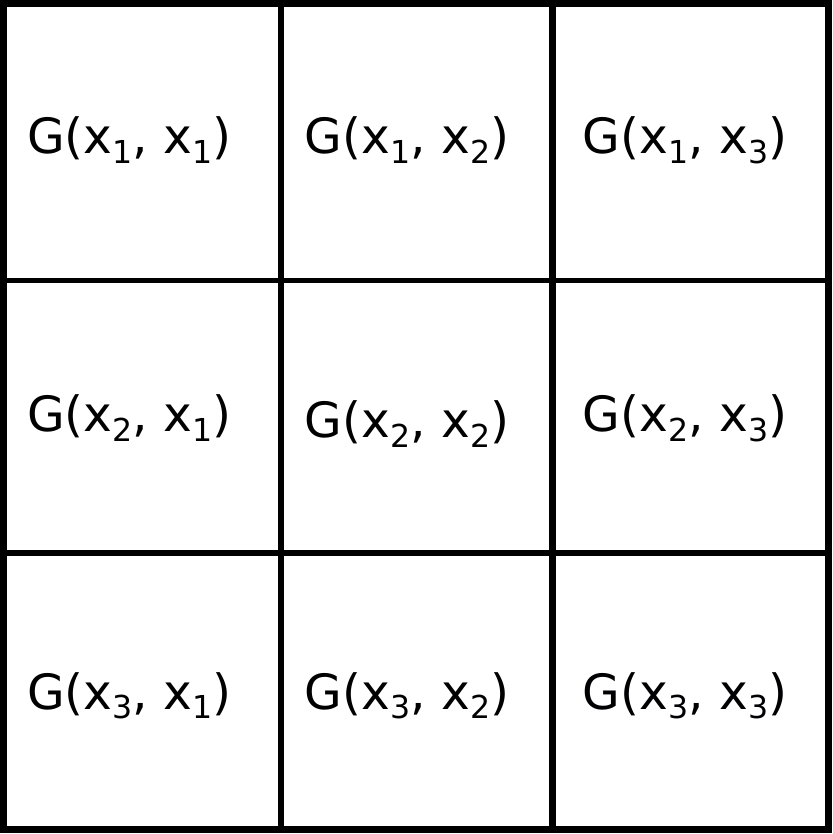}   
    \caption{Illustration of the components of the Gram matrix for $\dmt$=3. On the diagonal blocks are the Gram matrix of the frames, which are identical to the ones of the static texture model from \cite{gatys_texture_2015}. The other blocks contain the correlations between the adjacent frames.}       
    \label{fig:gram}
\end{figure}

Our model directly extends the static CNN texture model of Gatys et al. \cite{gatys_texture_2015}. In order to model a dynamic texture, we compute a set of spatio-temporal summary statistics from a given example movie of that texture. While the static texture model from \cite{gatys_texture_2015} only captures spatial summary statistics of a single image, our model additionally includes temporal correlations over several video frames.

We start with a given example video texture $\bX$ consisting of T frames $\bx_t$, for $t \in \{1,2,... T\}$. For each frame we compute the feature maps $\bF_{\ell}(\bx_t)$ in layer $\ell$ of a pre-trained CNN. Each column of $\bF_\ell(\bx_t)$ is a vectorised feature map and thus $\bF_\ell(\bx_t) \in \mathcal{R}^{M_\ell(\bx_t)\times N_\ell}$ where $N_\ell$ is the number of feature maps in layer $\ell$ and $M_\ell(\bx_t) = H_\ell(\bx_t)\times W_\ell(\bx_t)$ is the product of height and width of each feature map.

In the static texture model from \cite{gatys_texture_2015}, a texture is described by a set of Gram Matrices computed from the feature responses of the layers included in the texture model. A Gram Matrix from the feature maps in layer $\ell$ in response to image $\bx$ is defined as $\bG_\ell(\bx)=\frac{1}{M_\ell(\bx)}\bF_\ell(\bx)^\top\bF_\ell(\bx)$. 

To include temporal dependencies in our dynamic texture model we combine the feature maps of \dt consecutive frames and compute one large Gram Matrix from them (Fig.\ref{fig:gram}). We first concatenate the feature maps from the \dt frames along the second axis: 
$\bF_{\ell, \dmt}(\bx_1,\bx_2,...,\bx_{\dmt}) = \left[\bF_{\ell}(\bx_1),\bF_{\ell}(\bx_2),...,\bF_{\ell}(\bx_{\dmt})\right]$ such that $\bF_{\ell, \dmt}\in \mathcal{R}^{M_\ell\times \dmt N_\ell}$. Then we use this large feature matrix to compute a Gram Matrix $\bG_{\ell, \dmt}=\frac{1}{M_\ell}\bF_{\ell, \dmt}^\top\bF_{\ell, \dmt}$ that now also captures temporal dependencies of the order \dt (Fig.\ref{fig:gram}). Finally this Gram Matrix is averaged over all time windows $\dmt_i \text{ for } i \in [1,T-(\dmt-1)]$. Thus our model describes a dynamic texture by the spatio-temporal summary statistics
\begin{align}
  \bG_{\ell, \dmt}(\bX)  = \frac{1}{M_\ell}\sum_{i=1}^{T-(\dmt-1)}{\bF_{\ell, \dmt_i}^\top\bF_{\ell, \dmt_i}}
\end{align}
computed at all layers $\ell$ included in the model.
Compared to the static texture model \cite{gatys_texture_2015} this increases the number of parameters by a factor of $\dmt^2$.

\section{Texture generation}\label{sec:synthesis}
After extracting the spatio-temporal summary statistics from an example movie they can be used to generate new samples of the video texture. To that end we sequentially generate frames that match the extracted summary statistics. Each frame is generated by a gradient based pre-image search that starts from a white noise image to find an image that matches the texture statistics of the original video. 

Thus, to synthesise a frame $\hat{\bx}_t$ given the previous frames $[\hat{\bx}_{t-\dmt + 1},...,\hat{\bx}_{t-1}]$ we minimise the following loss function with respect to $\hat{\bx}_{t}$:
\begin{align}
\label{eq:opt}
\mathcal{L} &= \sum_\ell w_\ell E_\ell(\hat{\bx}_{t}) \\
E_\ell(\hat{\bx}_{t}) &= \frac{1}{4N_\ell^2}\sum_{ij}{\left(\bG_{\ell, \dmt + 1}(\hat{\bx}_{t-\dmt},...,\hat{\bx}_{t}) - \bG_{\ell, \dmt}(\bX)\right)_{ij}^{2}}
\end{align}

For all results presented here we included the layers `conv1\_1', `conv2\_1', `conv3\_1', `conv4\_1' and `conv5\_1' of the VGG-19 network \cite{simonyan_very_2014} in the texture model and weighted them equally ($w_l = w$).

The initial \dt -1 frames can be taken from the example movie, which allows the direct extrapolation of an existing video. Alternatively they can be generated jointly by starting with \dt randomly initialised frames and minimising $\mathcal{L}$ jointly with respect to $\hat{\bx}_{1}, \hat{\bx}_{2},...,\hat{\bx}_{\dmt}$.

In general this procedure can generate movies of arbitrary length because the extracted spatio-temporal summary statistics naturally do not depend on the length of the source video.

\section{Experiments and Results}

\begin{figure}
    \includegraphics[width=\linewidth]{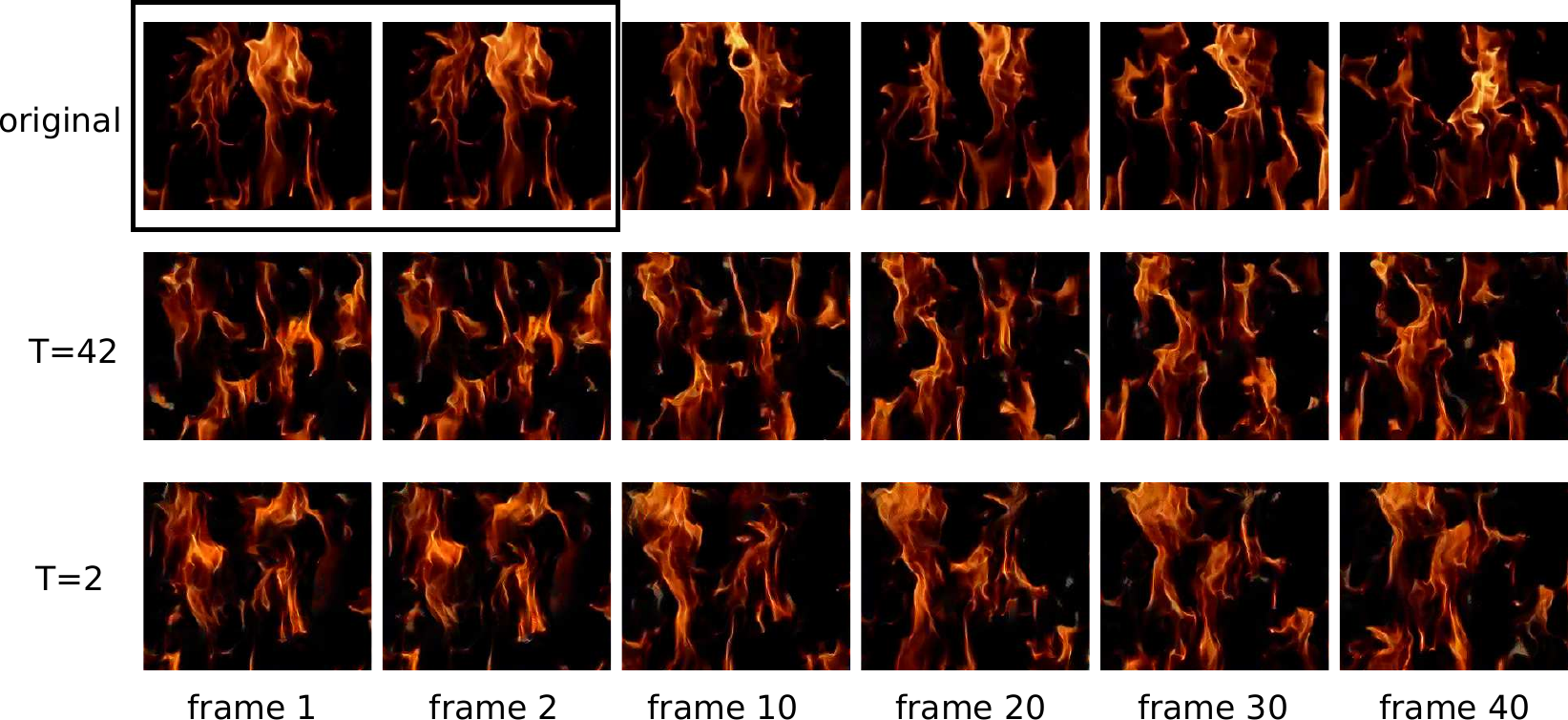}  \vspace{0.5cm}
    
    \includegraphics[width=\linewidth]{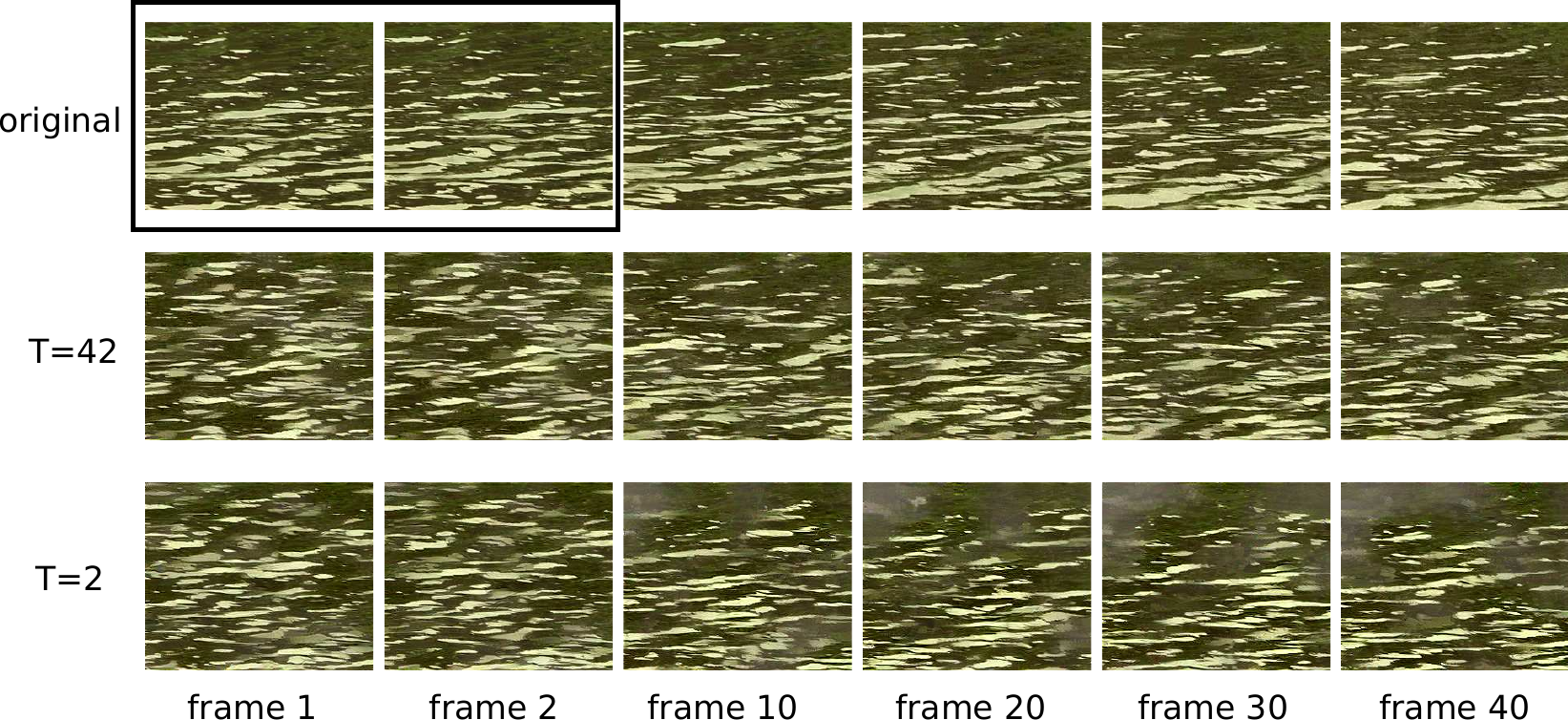}   

    \caption{Examples of generated video textures for $\dmt=2$ and two example textures. In the top rows frames of the original video are shown. For the frames in the middle rows, 42 original frames were used. For the frames in the bottom rows two original frames were used (the ones in the black box). The full videos can be found at \url{https://bethgelab.org/media/uploads/dynamic_textures/figure2/}.}       
    \label{fig:Fireplace}
\end{figure}

\begin{figure}
    \includegraphics[width=\linewidth]{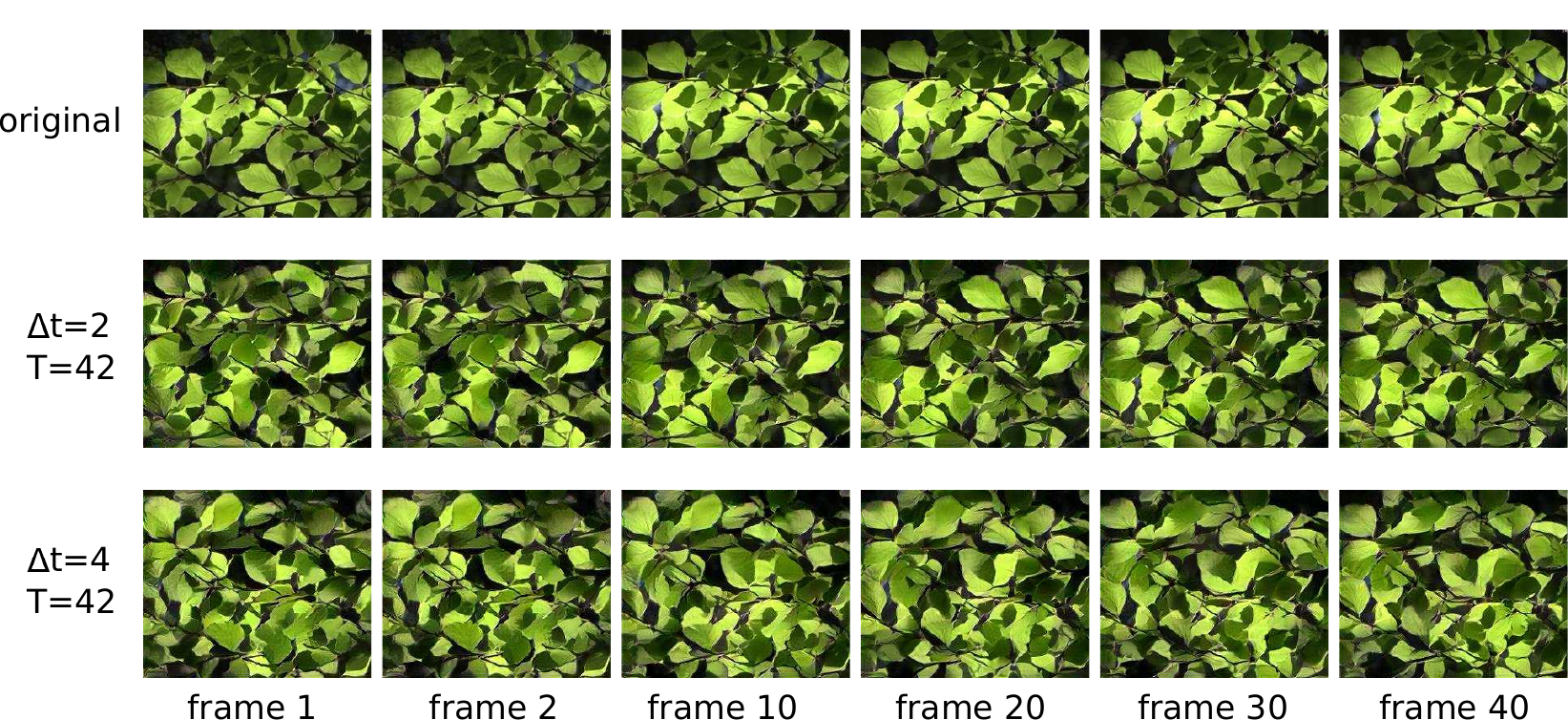}   \vspace{0.5cm}

    \includegraphics[width=\linewidth]{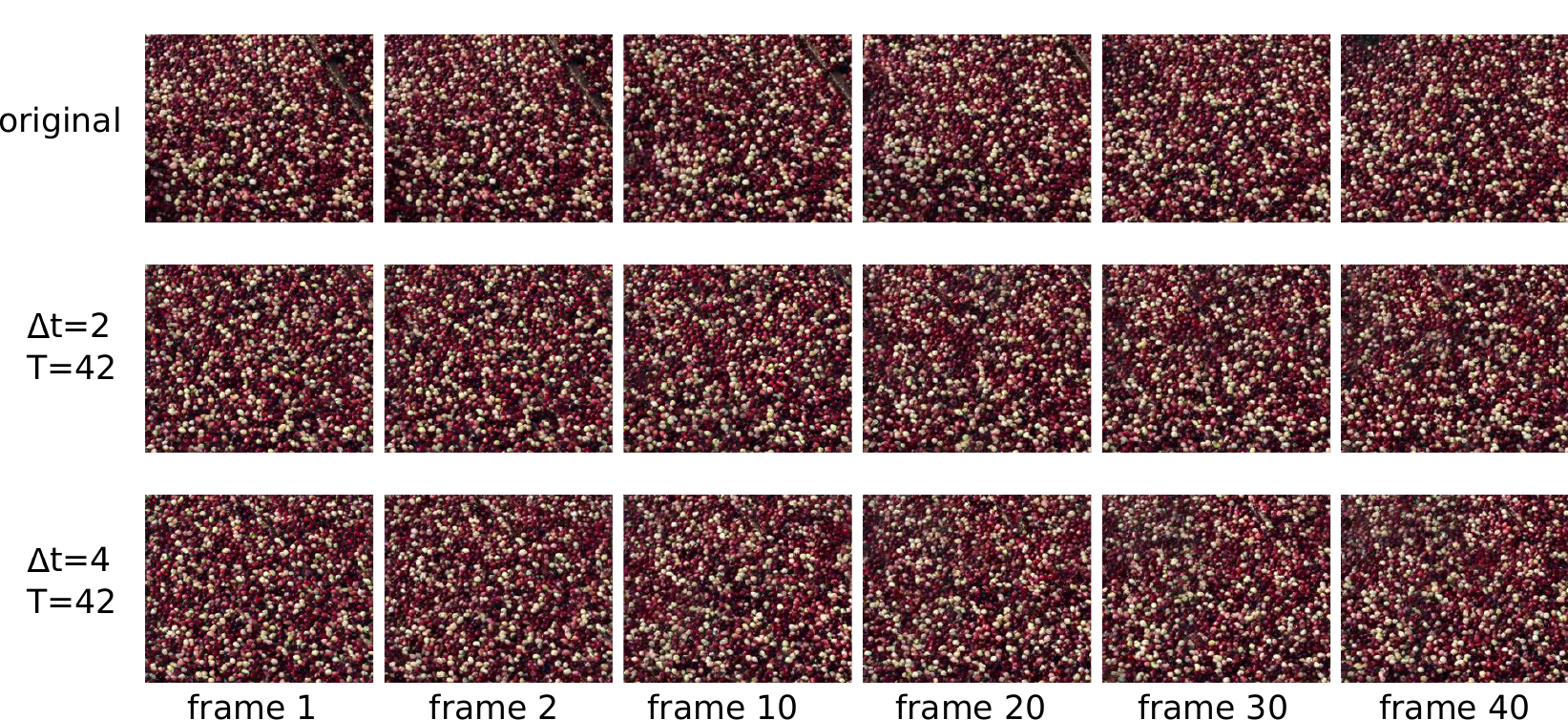}   
    
    \caption{Examples of generated videos for $\dmt=2$ (middle rows) and $\dmt=4$ (bottom rows). In the top rows frames of the original video are shown. 42 original frames were used. The global structure of the motion is not preserved. The full videos can be found at \url{https://bethgelab.org/media/uploads/dynamic_textures/figure3/}}       
    \label{fig:number_of_frames}
\end{figure}

Here we present dynamic textures generated by our model. We used example video textures from the DynTex database \cite{peteri_dyntex:_2010} and from the Internet. Each frame was generated by minimising the loss function for 500 iterations of the L-BFGS algorighm \cite{zhu_algorithm_1997}. All source textures and generated results can be found at \url{https://bethgelab.org/media/uploads/dynamic_textures/}.

First we show the results for $\dmt=2$ and random initialisation of the initial frames (Fig. \ref{fig:Fireplace}). We extracted the texture parameters from either $T=42$ frames of the source movie or just from a pair of frames $T=2$. Surprisingly we find that extracting the texture parameters from only two frames is often sufficient to generate diverse dynamic textures of arbitrary length (Fig. \ref{fig:Fireplace}, bottom rows).
However, the entropy of the generated frames is clearly higher for $T=42$ and for some videos (example: water) greyish regions appear in the generated texture if only two original frames are used.\\

Next we explored the effect of increasing the size of the time window \dt. Here we show results for $\dmt=2$ and $\dmt=4$. In general we noted that for most video textures varying the size of the time window \dt has little effect. We observed differences, however, in cases where the motion is more structured. 
For example, given a movie of a branch of a tree moving in the wind (Fig. \ref{fig:number_of_frames}, top row), the leaves are only moving slightly up and down for $\dmt=2$ (Fig. \ref{fig:number_of_frames}, middle row), whereas for $\dmt=4$ the motion extends over a larger range (Fig. \ref{fig:number_of_frames}, bottom row).

Still, even for $\dmt=4$, the generated video fails to capture the motion of the original texture. In particular, it fails to reproduce the global coherence of the motion in the source video. While in the source video, all leaves move together with the branch up or down, in the synthesised one some leaves move up while some move down at the same time. The disability to capture the global structure of the motion is even more apparent in the second example in Fig. \ref{fig:number_of_frames} and illustrates a limitation of our model.

\begin{figure}
    \includegraphics[width=\linewidth]{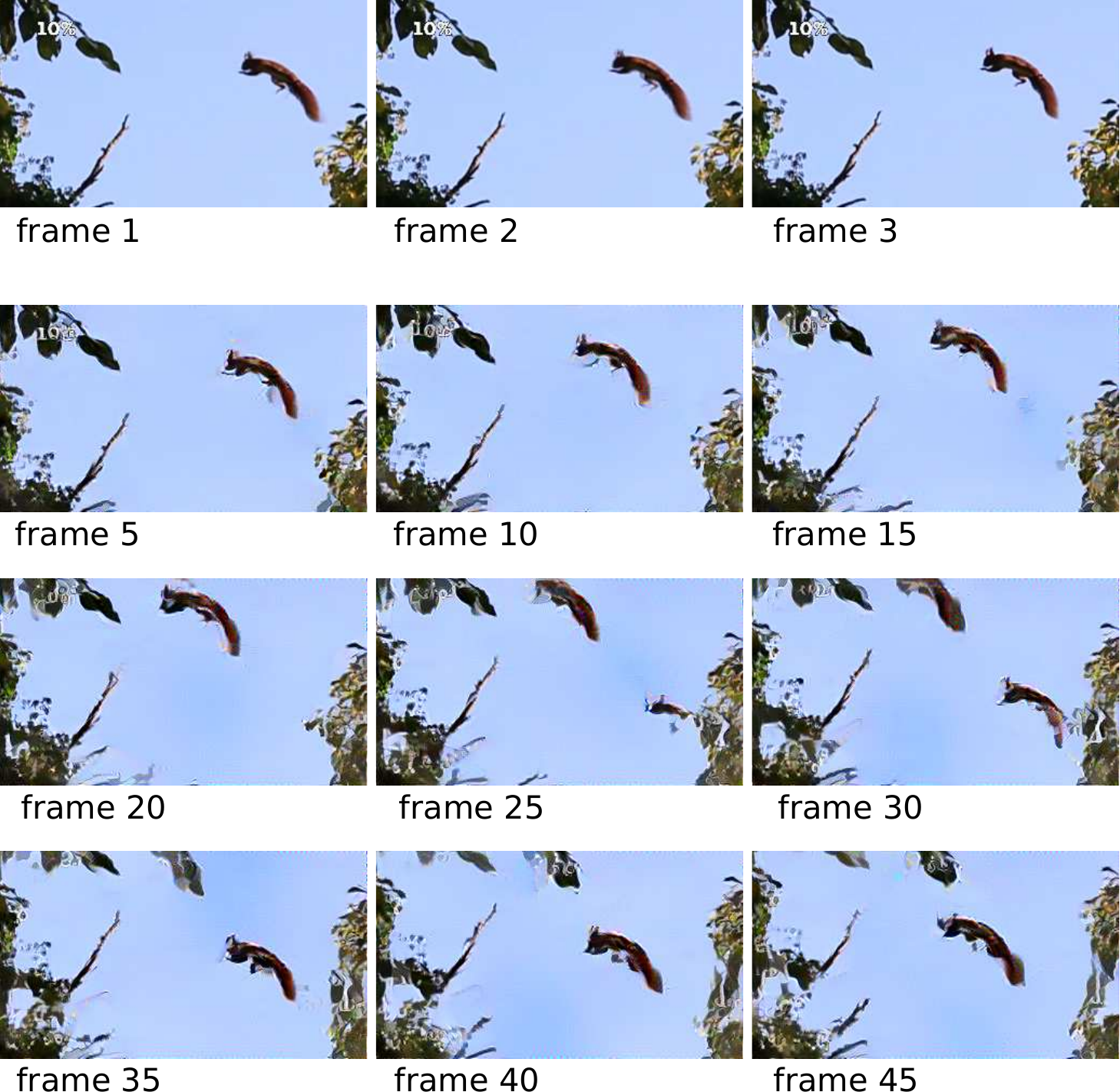}   
     \caption{Initialisation of the new video with the original frames. The first three frames shown are the original frames, the others are generated by our model. The full video can be found at \url{https://bethgelab.org/media/uploads/dynamic_textures/figure4/}.}
    \label{fig:squirrel}
\end{figure}

Finally, instead of generating a video texture from a random initialisation, we can also initialise with $\dmt -1$ frames from the example movie. In that way the spatial arrangement is kept and we are predicting the next frames of the movie based on the initial motion. We use three frames of the original video were to define the texture statistics ($\dmt=3$, $T=3$) (Fig. \ref{fig:squirrel}). The first two frames of the new movie are taken from the example and the following frames were sequentially generated as described in section \ref{sec:synthesis}. In the resulting video the different elements keep moving in the same direction: The squirrel continues flying to the top left, while the plants move upwards. If an element disappears from the image, it reappears somewhere else in the image. The generated movie can be arbitrarily long. In this case we used only the initial 3 frames to generate over 600 frames of a squirrel flying through the image and did not observe a decrease in image quality.

\section{Discussion}
We introduced a parametric model for dynamic textures based on the feature representations of a CNN trained on object recognition \cite{simonyan_very_2014}. In contrast to the CNN-based dynamic texture model by Xie et al. \cite{xie_synthesizing_2016}, our model can capture a large range of dynamic textures without the need to re-train the network for every given input texture.

Surprisingly we find that even when the temporal dependencies are extracted from as little as two adjacent frames our model still produces diverse looking dynamic textures (Fig. \ref{fig:Fireplace}). This is also true for non-texture movies with simple motion. We see that in this case we can generate a theoretically infinite movie repeating the same motion (Fig. \ref{fig:squirrel}.).

However, our model fails to capture structured motion with more complex temporal dependencies (Fig. \ref{fig:number_of_frames}). Possibly spatio-temporal CNN features or the inclusion of optical flow measures \cite{ruder_artistic_2016} might help to model temporal dependencies of that kind.  

In general though we find that for many dynamic textures the temporal statistics can be captured by second order dependencies between complex spatial features leading to a simple yet powerful parametric model for dynamic textures.

\bibliography{main}
\bibliographystyle{ieeetr}

\end{document}